\pdfoutput=1

\documentclass[11pt]{article}
\usepackage{acl}

\usepackage{times}
\usepackage{latexsym}
\usepackage{adjustbox}
\usepackage[most]{tcolorbox}

\usepackage[T1]{fontenc}

\usepackage[utf8]{inputenc}

\usepackage{microtype}

\usepackage{danudefs}
\usepackage{algorithmic}
\usepackage{algorithm}
\usepackage{color}
\usepackage{booktabs}
\usepackage{xspace}
\usepackage{url}
\usepackage{mathtools}
\usepackage{microtype}

\usepackage[english]{babel}
\usepackage{hyperref}
\addto\captionsenglish{%
}
\addto\extrasenglish{%
}

\usepackage{todonotes}

\makeatletter
\if@todonotes@disabled

\else

\fi
\makeatother

\usepackage{xifthen}

\usepackage{arydshln}
\makeatletter
\def\adl@drawiv#1#2#3{%
        \hskip.5\tabcolsep
        \xleaders#3{#2.5\@tempdimb #1{1}#2.5\@tempdimb}%
                #2\z@ plus1fil minus1fil\relax
        \hskip.5\tabcolsep}
\newcommand{\cdashlinelr}[1]{%
  \noalign{\vskip\aboverulesep
          \global\let\@dashdrawstore\adl@draw
          \global\let\adl@draw\adl@drawiv}
  \cdashline{#1}
  \noalign{\global\let\adl@draw\@dashdrawstore
          \vskip\belowrulesep}}
\makeatother
\DeclareMathOperator{\mean}{\mathrm{mean}}

\title{\emph{Sense Embeddings are also Biased} -- Evaluating Social Biases in \\Static and Contextualised Sense Embeddings}

\author{Yi Zhou$^{1}$ \And Masahiro Kaneko$^{2}$ \\ 
$\text{University of Liverpool}^{1}$, \ $\text{Tokyo Institute of Technology}^{2}$, \ $\text{Amazon}^{3}$  \\
\texttt{\{y.zhou71,danushka\}@liverpool.ac.uk} \\
\texttt{masahiro.kaneko@nlp.c.titech.ac.jp} 
\And Danushka Bollegala$^{1,3}$\thanks{\ \ Danushka Bollegala holds concurrent appointments as a Professor at University of Liverpool and as an Amazon Scholar. This paper describes work performed at the University of Liverpool and is not associated with Amazon.} \\
}

\date{}

\begin{document}
\maketitle

\begin{abstract}
Sense embedding learning methods learn different embeddings for the different senses of an ambiguous word.
One sense of an ambiguous word might be socially biased while its other senses remain unbiased.
In comparison to the numerous prior work evaluating the social biases in pretrained word embeddings, the biases in sense embeddings have been relatively understudied.
We create a benchmark dataset for evaluating the social biases in sense embeddings and propose novel sense-specific bias evaluation measures.
We conduct an extensive evaluation of multiple static and contextualised sense embeddings for various types of social biases using the proposed measures.
Our experimental results show that even in cases where no biases are found at word-level, there still exist worrying levels of social biases at sense-level, which are often ignored by the word-level bias evaluation measures.\footnote{The dataset and evaluation scripts are available at \url{github.com/LivNLP/bias-sense}.}
\end{abstract}

\section{Introduction}
\label{sec:intro}

Sense embedding learning methods use different vectors to represent the different senses of an ambiguous word~\cite{reisinger2010multi,neelakantan2014efficient,loureiro2019language}.
Although numerous prior works have studied social biases in static and contextualised word embeddings, social biases in sense embeddings remain underexplored~\cite{kaneko-bollegala-2019-gender,Kaneko:EACL:2021b,Kaneko:EACL:2021b,DBLP:conf/acl/RavfogelEGTG20,Dev:2019,schick2020self,wang2020doublehard}.  

We follow \newcite{shah-etal-2020-predictive} and define social biases to be \emph{predictive biases with respect to protected attributes} made by NLP systems.
Even if a word embedding is unbiased, some of its senses could still be associated with unfair social biases.
For example, consider the ambiguous word \emph{black}, which has two adjectival senses according to the WordNet~\cite{miller1998wordnet}: (1) black as a \emph{colour} (\emph{being of the achromatic colour of maximum darkness}, sense-key=\textbf{black\%3:00:01}) and (2) black as a \emph{race} (\emph{of or belonging to a racial group especially of sub-Saharan African origin}, sense-key=\textbf{black\%3:00:02}).
However, only the second sense of \emph{black} is often associated with racial biases.

\begin{figure}[t]
\centering
\includegraphics[width=0.48\textwidth]{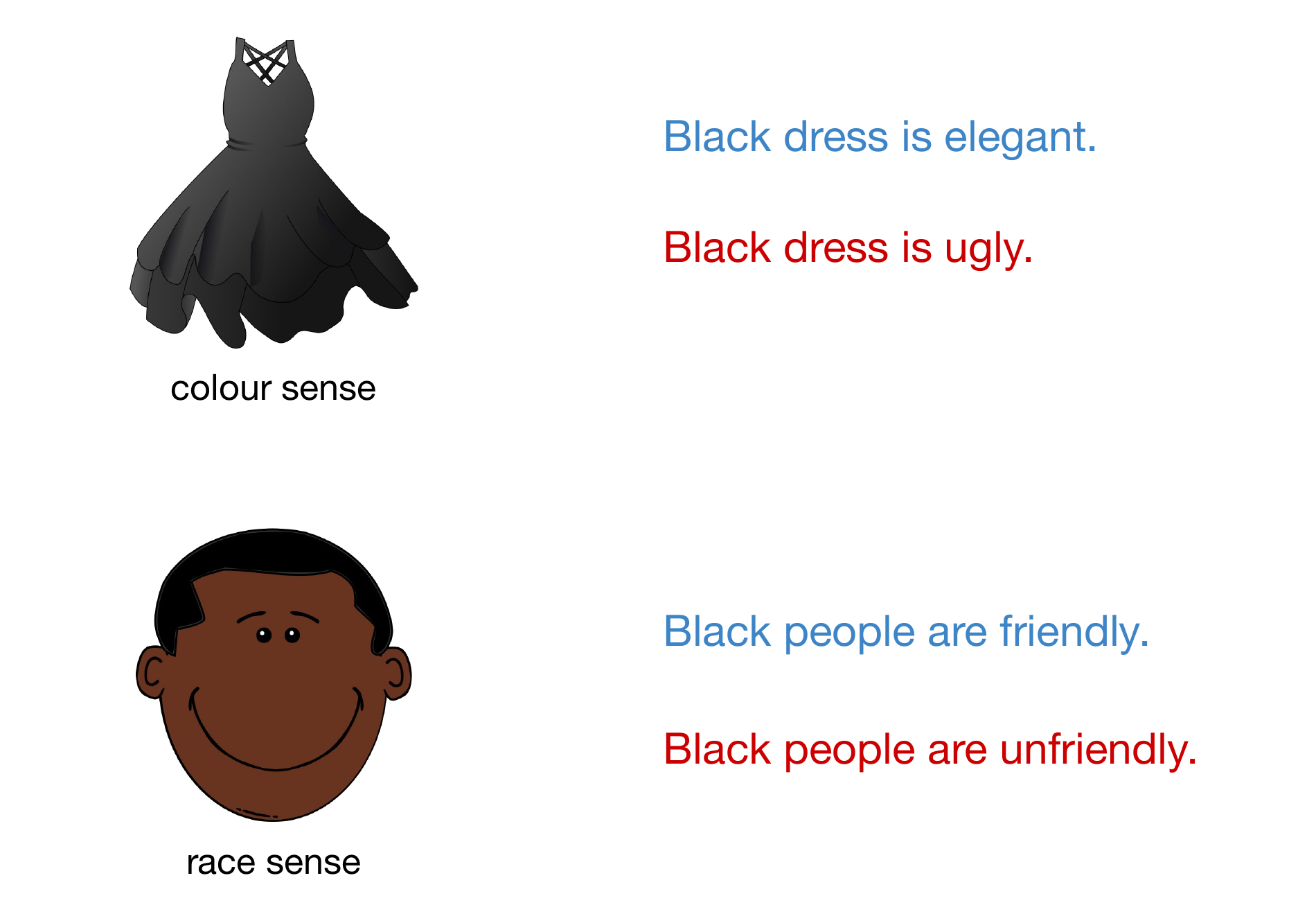}
\caption{Example sentences from the Sense-Sensitive Social Bias dataset for the two senses of the ambigous word \emph{black}.  Top two sentences correspond to the colour sense of black, whereas the bottom two sentences correspond to its racial sense.
Stereotypical examples that associate a sense with an unpleasant attribute are shown in red, whereas anti-stereotypical examples that associate a sense with a pleasant attribute are shown in blue.}
\label{fig:teaser-fig}
\end{figure}

Owing to (a) the lack of evaluation benchmarks for the social biases in sense embeddings, 
and (b) not being clear how to extend the bias evaluation methods that are proposed for static and contextualised embeddings to evaluate social biases in sense embeddings, existing social bias evaluation datasets and metrics do not consider multiple senses of words, thus not suitable for evaluating biases in sense embeddings.

To address this gap, we evaluate social biases in state-of-the-art (SoTA) static sense embeddings such as LMMS~\cite{loureiro2019language} and ARES~\cite{scarlini2020more}, as well as contextualised sense embeddings obtained from SenseBERT~\cite{levine-etal-2020-sensebert}.
To the best of our knowledge, we are the first to conduct a systematic evaluation of social biases in sense embeddings.
Specifically, we make two main contributions in this paper:
\begin{itemize}
    \item First, to evaluate social biases in static sense embeddings, we extend previously proposed benchmarks for evaluating social biases in static (sense-insensitive) word embeddings by manually assigning sense ids to the words considering their social bias types expressed in those datasets (\autoref{sec:static}).
    \item Second, to evaluate social biases in sense-sensitive contextualised embeddings, we create the Sense-Sensitive Social Bias (\textbf{SSSB}) dataset, a novel template-based dataset containing sentences annotated for multiple senses of an ambiguous word considering its stereotypical social biases (\autoref{sec:contextualised}). 
    An example from the SSSB dataset is shown in \autoref{fig:teaser-fig}.
\end{itemize}

Our experiments show that, similar to word embeddings, both static as well as contextualised sense embeddings also encode worrying levels of social biases. 
Using SSSB, we show that the proposed bias evaluation measures for sense embeddings capture different types of social biases encoded in existing SoTA sense embeddings.
More importantly, we see that even when social biases cannot be observed at word-level, such biases are still prominent at sense-level, raising concerns on existing evaluations that consider only word-level social biases.

\section{Related Work}
\label{sec:related}

Our focus in this paper is the evaluation of social biases in English and \emph{not} the debiasing methods.
We defer the analysis for languages other than English and developing debiasing methods for sense embeddings to future work.
Hence, we limit the discussion here only to bias evaluation methods.
\noindent
\paragraph{Biases in Static Embeddings:}
The Word Embedding Association Test~\cite[\textbf{WEAT};][]{WEAT} evaluates the association between two sets of target concepts (e.g. \emph{male} vs.~\emph{female}) and two sets of attributes (e.g. Pleasant (\emph{love, cheer}, etc.) vs.~Unpleasant (\emph{ugly, evil}, etc.)).
Here, the association is measured using the cosine similarity between the word embeddings.
\newcite{ethayarajh-etal-2019-understanding} showed that WEAT systematically overestimates the social biases and proposed relational inner-product association (\textbf{RIPA}), a subspace projection method, to overcome this problem.

Word Association Test~\cite[\textbf{WAT};][]{du-etal-2019-exploring} calculates a gender information vector for each word in an association graph~\cite{Deyne2019TheW} by propagating information related to masculine and feminine words.
Additionally, word analogies are used to evaluate gender bias in static embeddings \cite{Tolga:NIPS:2016,manzini-etal-2019-black,Zhao:2018ab}. 
\newcite{loureiro2019language} showed specific examples of gender bias in static sense embeddings.
However, these datasets do not consider word senses, hence are unfit for evaluating social biases in sense embeddings.

\noindent
\paragraph{Biases in Contextualised Embeddings:}
\newcite{may-etal-2019-measuring} extended WEAT to sentence encoders by creating artificial sentences using templates and used cosine similarity between the sentence embeddings as the association metric.
\newcite{kurita-etal-2019-measuring} proposed the log-odds of the target and prior probabilities of the sentences computed by masking respectively only the target vs.~both target and attribute words.
Template-based approaches for generating example sentences for evaluating social biases do not require human annotators to write examples, which is often slow, costly and require careful curation efforts.
However, the number of sentence patterns that can be covered via templates is often small and less diverse compared to manually written example sentences.

To address this drawback, \newcite[\textbf{StereoSet};][]{Nadeem:2021} created human annotated contexts of social bias types, while 
\newcite{crows-pairs} proposed Crowdsourced Stereotype Pairs benchmark (\textbf{CrowS-Pairs}).
Following these prior work, we define a stereotype as a commonly-held association between a group and some attribute.
These benchmarks use sentence pairs of the form ``\emph{She is a nurse/doctor}''.
StereoSet calculates log-odds by masking the modified tokens (\emph{nurse}, \emph{doctor}) in a sentence pair, whereas CrowS-Pairs calculates log-odds by masking their unmodified tokens (\emph{She}, \emph{is}, \emph{a}).

\newcite{kaneko2021unmasking} proposed All Unmasked Likelihood (\textbf{AUL}) and AUL with Attention weights (\textbf{AULA}), which calculate log-likelihood by predicting all tokens in a test case, given the contextualised embedding of the unmasked input.

\section{Evaluation Metrics for Social Biases in Static Sense Embeddings}
\label{sec:static}


We extend the WEAT and WAT datasets that have been frequently used in prior work for evaluating social biases in static word embeddings such that they can be used to evaluate sense embeddings.
These datasets compare the association between a target word $w$ and some (e.g. pleasant or unpleasant) attribute $a$, using the cosine similarity, $\cos(\vec{w},\vec{a})$, computed using the static word embeddings $\vec{w}$ and $\vec{a}$ of respectively $w$ and $a$.
Given two same-sized sets of \emph{target} words $\cX$ and $\cY$ and two sets of \emph{attribute} words $\cA$ and $\cB$,
the bias score, $s(\cX,\cY,\cA,\cB)$, for each target is calculated as follows:

{\small
\begin{align}
&s(\cX,\cY,\cA,\cB) = \sum_{\vec{x} \in \cX} w(\vec{x}, \cA, \cB) - \sum_{\vec{y} \in \cY} w(\vec{y}, \cA, \cB) \\
&w(\vec{t}, \cA, \cB) = \underset{\vec{a} \in \cA}{\mean} \cos(\vec{t}, \vec{a}) - \underset{\vec{b} \in \cB}{\mean} \cos(\vec{t}, \vec{b})
\end{align}
}

Here, $\cos(\vec{a}, \vec{b})$ is the cosine similarity\footnote{Alternatively, inner-products can be used to extend RIPA.} between the embeddings $\vec{a}$ and $\vec{b}$.
The one-sided $p$-value for the permutation test for $\cX$ and $\cY$ is calculated as the probability of $s(\cX_i,\cY_i,\cA,\cB) > s(\cX,\cY,\cA,\cB)$.
The effect size is calculated as the normalised measure given by \eqref{eq:effect}:
\begin{align}
\label{eq:effect}
\frac{\underset{x \in \cX}{\mean\ } w(x, \cA,\cB) - \underset{y \in \cY}{\mean\ } w(y, \cA, \cB)}{\underset{t \in \cX \cup \cY}{\mathrm{sd}} w(t, \cA, \cB)}
\end{align}

We repurpose these datasets for evaluating the social biases in \emph{sense} embeddings as follows.
For each target word in WEAT, we compare each sense $s_i$ of the target word $w$ against each sense $a_j$ of a word selected from the association graph using their corresponding sense embeddings, $\vec{s}_i, \vec{a}_j$, and use the maximum similarity over all pairwise combinations (i.e. $\max_{i,j} \cos(\vec{s}_i,\vec{a}_j)$) as the word association measure.
Measuring similarity between two words as the maximum similarity over all candidate senses of each word is based on the assumption that two words in a word-pair would mutually disambiguate each other in an association-based evaluation~\cite{Pilehvar:2019}, and has been used as a heuristic for disambiguating word senses~\cite{reisinger2010multi}.

WAT considers only gender bias and calculates the gender information vector for each word in a word association graph created with Small World of Words project~\cite{Deyne2019TheW} by propagating information related to masculine and feminine words $(w_m^i, w_f^i) \in \cL$ using a random walk approach~\cite{Zhou2003LearningWL}.
It is non-trivial to pre-specify the sense of a word in a large word association graph considering the paths followed by a random walk.
The gender information is encoded as a vector ($b_m$, $b_f$) in $2$ dimensions, where $b_m$ and $b_f$ denote the masculine and feminine orientations of a word, respectively.
The bias score of a word is defined as $\log(b_m / b_f)$.
The gender bias of word embeddings are evaluated using the Pearson correlation coefficient between the bias score of each word and the score given by \eqref{eq:bias-score}, computed as the average over the differences of cosine similarities between masculine and feminine words.
\begin{align}
\label{eq:bias-score}
\frac{1}{|\cL|} \sum_{i=1}^{|\cL|} \left( \cos(\vec{w}, \vec{w}_m^i) - \cos(\vec{w}, \vec{w}_f^i) \right)
\end{align}

To evaluate gender bias in sense embeddings, we follow the method that is used in WEAT, and take $\max_{i,j} \cos(\vec{s}_i,\vec{a}_j)$) as the word association measure.


\section{Sense-Sensitive Social Bias Dataset}


\begin{table}[t]
\centering
\small
\begin{adjustbox}{width=\linewidth}
\begin{tabular}{l c c c}
\toprule 
  Category              & noun vs. & race vs. & nationality vs. \\ 
                & verb     & colour        & language \\
\midrule        
\#pleasant words     & 14        & 5                 & 18 \\
\#unpleasant words    & 18        & 5                 & 15 \\
\#target words        & 6         & 1                 & 16 \\
\#templates     & 1         & 4                 & 4 \\
\#test cases    & 324       & 733               & 2304 \\
\bottomrule
\end{tabular}
\end{adjustbox}
\caption{Statistics of the the SSSB dataset.}
\label{tbl:data}
\end{table}

\begin{table*}[t!]
\begin{tabular}{l p{11.5cm}}
\toprule 
  Category  & Ambiguous words considered       \\ 
\midrule        
noun vs.~verb  & engineer, carpenter, guide, mentor, judge, nurse \\
race vs.~colour  & black \\
nationality vs.~language & Japanese, Chinese, English, Arabic, German, French, Spanish, Portuguese, Norwegian, Swedish, Polish, Romanian, Russian, Egyptian, Finnish, Vietnamese \\
\bottomrule
\end{tabular}
\caption{Bias categories covered in the SSSB dataset}
\label{tbl:composition}
\end{table*}

Contextualised embeddings such as the ones generated by masked language models (MLMs) return different vectors for the same word in different contexts.
However, the datasets discussed in \autoref{sec:static} do not provide contextual information for words and cannot be used to evaluate contextualised embeddings.
Moreover, the context in which an ambiguous word occurs determines its word sense.
Contextualised sense embedding methods such as SenseBERT (fine-tuned using WordNet super senses), have shown to capture word sense information in their contextualised embeddings~\cite{Zhou:PACLIC:2021}.

CrowS-Pairs and StereoSet datasets were proposed for evaluating contextualised word embeddings.
Specifically, an MLM is considered to be unfairly biased if it assigns higher pseudo log-likelihood scores for stereotypical sentences, $S^{\rm st}$, than anti-stereotypical ones, $S^{\rm at}$. 
However, both of those datasets do not consider multiple senses of words and cannot be used to evaluate social biases in contextualised sense embeddings.

To address this problem, we create the  Sense-Sensitive Social Bias (SSSB)  dataset, containing template-generated sentences covering multiple senses of ambiguous words for three types of social biases: \emph{gender}, \emph{race} and \emph{nationality}.
Templates are used in the same sense as in prior work such as \newcite{kurita-etal-2019-measuring}.
For example, we manually create templates such as 
[gender word] is a [pleasant/unpleasant attribute] engineer. 
We then fill the gender word by male and female gender pronouns (he/she), pleasant attributes (e.g. careful, skilful, efficient, etc.) and unpleasant attributes (e.g. clumsy, unskillful, inefficient, etc.) to generate many example sentences demonstrating social biases.

To the best of our knowledge, SSSB is the first-ever dataset created for the purpose of evaluating social biases in sense embeddings.
\autoref{tbl:data} shows the summary statistics of the SSSB dataset.
\autoref{tbl:composition} shows the bias categories covered in the SSSB dataset. 
Next, we describe the social biases covered in this dataset.

\subsection{Nationality vs. Language Bias}

These examples cover social biases related to a nationality (racial) or a language (non-racial). 
Each test case covers two distinct senses and the following example shows how they represent biases.
\emph{Japanese people are nice} is an anti-stereotype for \emph{Japanese} as a nationality because it is associated with a pleasant attribute (i.e. \emph{nice}) in this example sentence.
On the other hand, \emph{Japanese people are stupid} is a stereotype for \emph{Japanese} as a nationality because it is associated with an unpleasant attribute (i.e. \emph{stupid}).
These can be considered as examples of racial biases.

Likewise, for the language sense of Japanese we create examples as follows.
\emph{Japanese language is difficult to understand} is a stereotype for \emph{Japanese} as a language because it is associated with an unpleasant attribute (i.e. \emph{difficult}).
On the other hand, \emph{Japanese language is easy to understand} is an anti-stereotype for \emph{Japanese} as a language because it is associated with a pleasant attribute (i.e. \emph{easy}).

In SSSB, we indicate the sense-type, WordNet sense-id and the type of social bias in each example as follows: 
\begin{tcolorbox}[colback=black!5!white,colframe=black!75!black]
\emph{Japanese people are beautiful.} \newline [nationality, japanese\%1:18:00::, anti]
\end{tcolorbox}
Here, sense-type is nationality, sense-id as specified in the WordNet is \emph{japanese\%1:18:00::} and the bias is anti (we use the labels \emph{anti} and \emph{stereo} to denote respectively anti-stereotypical and stereotypical biases). 

We use the likelihood scores returned by an MLM to nationality vs.~language sentence pairs as described further in \autoref{sec:contextualised} to evaluate social biases in MLMs.
Essentially, if the likelihood score returned by an MLM for the example that uses an unpleasant attribute is higher than the one that uses a pleasant attribute for a member in the disadvantaged group, then we consider the MLM to be socially biased.
Moreover, if a member in the disadvantaged group is associated with a positive attribute in a stereotypical manner, we consider this as a anti-stereotype case. 
For example, we classify \emph{Asians are smart} as anti-stereotype rather than ``positive'' stereotypes following prior work on word-level or sentence-level bias evaluation datasets (e.g., Crows-Pairs and StereoSet) to focus on more adverse types of biases that are more direct and result in discriminatory decisions against the disadvantaged groups.

Note that one could drop the modifiers such as \emph{people} and \emph{language} and simplify these examples such as \emph{Japanese are nice} and \emph{Japanese is difficult} to generate additional test cases. 
However, the sense-sensitive embedding methods might find it difficult to automatically disambiguate the correct senses without the modifiers such as \emph{language} or \emph{people}.
Therefore, we always include these modifiers when creating examples for nationality vs.~language bias in the SSSB dataset.

\subsection{Race vs. Colour Bias}
The word \emph{black} can be used to represent the race (black people) or the colour.
We create examples that distinguish these two senses of black as in the following example.
\emph{Black people are friendly} represents an anti-stereotype towards \emph{black} because it is associated with a pleasant attribute (i.e. \emph{friendly}) of a disadvantaged group whereas, \emph{Black people are arrogant} represents a stereotype because it is associated with an unpleasant attribute (i.e. \emph{arrogant}).

On the other hand, for the colour black, \emph{The black dress is elegant} represents an anti-stereotype because it is associated with a pleasant attribute (i.e. \emph{elegant}), whereas \emph{The black dress is ugly} represents a stereotype because it is associated with an unpleasant attribute (i.e. \emph{ugly}).
If the likelihood score returned by an MLM for a sentence containing the racial sense with an unpleasant attribute is higher than one that uses a pleasant attribute, the MLM is considered to be socially biased.

\subsection{Gender Bias in Noun vs. Verb Senses}
\label{sec:noun-vs-verb}

To create sense-related bias examples for gender,\footnote{We consider only male and female genders in this work} we create examples based on occupations. 
In particular, we consider the six occupations: \emph{engineer}, \emph{nurse}, \emph{judge}, \emph{mentor}, \emph{(tour) guide}, and \emph{carpenter}.
These words can be used in a noun sense (e.g. \emph{engineer is a person who uses scientific knowledge to solve practical problems}, \emph{nurse is a person who looks after patients}, etc.) as well as in a verb sense expressing the action performed by a person holding the occupation (e.g. \emph{design something as an engineer}, \emph{nurse a baby}, etc.).
Note that the ambiguity here is in the occupation (noun) vs.~action (verb) senses and not in the gender, whereas the bias is associated with the gender of the person holding the occupation.

To illustrate this point further, consider the following examples.
\emph{She is a talented engineer} is considered as an anti-stereotypical example for the noun sense of \emph{engineer} because females (here considered as the disadvantaged group) are not usually associated with pleasant attributes (i.e. \emph{talented}) with respect to this occupation (i.e. \emph{engineer}).
\emph{He is a talented engineer} is considered as a stereotypical example for the noun sense of engineer because males (here considered as the advantaged group) are usually associated with pleasant attributes with regard to this occupation.
As described in \autoref{sec:contextualised}, if an MLM assigns a higher likelihood to the stereotypical example (second sentence) than the anti-stereotypical example (first sentence), then that MLM is considered to be gender biased.

On the other hand, \emph{She is a clumsy engineer} is considered to be a stereotypical example for the noun sense of engineer because females (i.e. disadvantaged group) are historically associated with such unpleasant attributes (i.e. \emph{clumsy}) with respect to such male-dominated occupations.
Likewise, \emph{He is a clumsy engineer} is considered as an anti-stereotypical example for the noun sense of engineer because males (i.e. advantaged group) are not usually associated with such unpleasant attributes (i.e. \emph{clumsy}).
Here again, if an MLM assigns a higher likelihood to the stereotypical example (first sentence) than the anti-stereotypical example (second sentence), then it is considered to be gender biased. 
Note that the evaluation direction with respect to male vs.~female pronouns used in these examples is opposite to that in the previous paragraph because we are using an unpleasant attribute in the second set of examples.

Verb senses are also used in the sentences that contain gender pronouns in SSSB.
For example, for the verb sense of \emph{engineer}, we create examples as follows:
\emph{She used novel material to engineer the bridge}.
Here, the word engineer is used in the verb sense in a sentence where the subject is a female.
The male version of this example is as follows:
\emph{He used novel material to engineer the bridge}.
In this example, a perfectly unbiased MLM should not systematically prefer one sentence over the other between the two sentences both expressing the verb sense of the word \emph{engineer}.

\section{Evaluation Metrics for Social Biases in Contextualised Sense Embeddings}
\label{sec:contextualised}

For a contextualised (word/sense) embedding under evaluation, we compare its pseudo-likelihood scores for stereotypical and anti-stereotypical sentences for each sense of a word in SSSB, using 
AUL~\cite{kaneko2021unmasking}.\footnote{The attention-weighted variant (AULA) is not used because contextualised sense embeddings have different structures of attention from contextualised embeddings, and it is not obvious which attention to use in the evaluations.}
AUL is known to be robust against the frequency biases of words and provides more reliable estimates compared to the other metrics for evaluating social biases in MLMs.
Following the standard evaluation protocol, we provide AUL the complete sentence $S = w_1, \ldots, w_{|S|}$, which contains a length $|S|$ sequence of tokens $w_i$, to an MLM with pretrained parameters $\theta$.
We first compute $\mathrm{PLL}(S)$, the Pseudo Log-Likelihood (PLL) for predicting all tokens in $S$ excluding begin and end of sentence tokens, given by~\eqref{eq:AUL}:
\begin{align}
    \label{eq:AUL}
    \mathrm{PLL}(S) \coloneqq \frac{1}{|S|} \sum_{i=1}^{|S|} \log P(w_i | S; \theta)
\end{align}

Here, $P(w_i | S; \theta)$ is the probability assigned by the MLM to token $w_i$ conditioned on $S$. 
The fraction of sentence-pairs in SSSB, where higher PLL scores are assigned to the stereotypical sentence than the anti-stereotypical one is considered as the AUL \emph{bias score} of the MLM associated with the contextualised embedding, and is given by \eqref{eq:score}:

{\small 
\begin{align}
    \label{eq:score}
  \mathrm{AUL} = \left( \frac{100}{N}\sum_{(S^{\rm st}, S^{\rm at})} \mathbb{I}(\mathrm{PLL}(S^{\rm st}) > \mathrm{PLL}(S^{\rm at}))\right) - 50
\end{align}
}

Here, $N$ is the total number of sentence-pairs in SSSB and $\mathbb{I}$ is the indicator function, which returns $1$ if its argument is True and $0$ otherwise.
AUL score given by \eqref{eq:score} falls within the range $[-50,50]$ and an unbiased embedding would return bias scores close to 0, whereas bias scores less than or greater than 0 indicate bias directions towards respectively the anti-stereotypical or stereotypical examples.

\section{Experiments}
\label{sec:exp}

\subsection{Bias in Static Embeddings}
\label{exp:static}

To evaluate biases in static sense embeddings, we select two current SoTA sense embeddings: LMMS\footnote{\url{https://github.com/danlou/LMMS}}~\citep{loureiro2019language} and ARES\footnote{\url{http://sensembert.org}}~\citep{scarlini2020more}.
In addition to WEAT and WAT datasets described in \autoref{sec:static}, we also use SSSB to evaluate static sense embeddings using the manually assigned sense ids for the target and attribute words, ignoring their co-occurring contexts.
LMMS and ARES sense embeddings associate each sense of a lexeme with a sense key and a vector, which we use to compute cosine similarities as described in \autoref{sec:static}.
To compare the biases in a static sense embedding against a corresponding sense-insensitive static word embedding version, we compute a static word embedding $\vec{w}$, for an ambiguous word $w$ by taking the average (\textbf{avg}) over the sense embeddings $\vec{s}_{i}$ for all of $w$'s word senses as given in \eqref{eq:word-embeddings-average}, where $M(w)$ is the total number of senses of $w$:
\begin{align}
    \vec{w}=\frac{\sum_i^{M(w)} \vec{s_i}}{M(w)}
    \label{eq:word-embeddings-average}
\end{align}

This would simulate the situation where the resultant embeddings are word-specific but not sense-specific, while still being comparable to the original sense embeddings in the same vector space.
As an alternative to \eqref{eq:word-embeddings-average}, which weights all different senses of $w$ equally, we can weight different senses by their frequency. 
However, such sense frequency statistics are not always available except for sense labelled corpora such as SemCor~\cite{SemCor}.
Therefore, we use the unweighted average given by \eqref{eq:word-embeddings-average}.

From \autoref{tbl:staic} we see that in WEAT\footnote{Three bias types (European vs.~African American, Male vs.~Female, and Old vs.~Young) had to be excluded because these biases are represented using personal names that are not covered by LMMS and ARES sense embeddings.} in all categories considered, sense embeddings always report a higher bias compared to their corresponding sense-insensitive word embeddings.
This shows that even if there are no biases at the word-level, we can still observe social biases at the sense-level in WEAT.
However, in the WAT dataset, which covers only gender-related biases, we see word embeddings to have higher biases than sense embeddings.
This indicates that in WAT gender bias is more likely to be observed in static word embeddings than in static sense embeddings.

In SSSB, word embeddings always report the same bias scores for the different senses of an ambiguous word because static word embeddings are neither sense nor context sensitive.
As aforementioned, the word ``black'' is bias-neutral with respect to the colour sense, while it often has a social bias for the racial sense.
Consequently, for \emph{black} we see a higher bias score for its racial  than colour sense in both LMMS and ARES sense embeddings.

\begin{table}[t]
\centering
\adjustbox{max width=\linewidth}{%
\begin{tabular}{lcc}
\toprule
& LMMS  & ARES \\
\cmidrule(lr){2-2} \cmidrule(lr){3-3}
Dataset & word/sense & word/sense \\
\midrule
\textbf{WEAT} & & \\
Flowers vs Insects & 1.63/\textbf{2.00} & 1.58/\textbf{2.00} \\
Instruments vs Weapons & 1.42/\textbf{2.00} & 1.37/\textbf{1.99} \\
Math vs Art & 1.52/\textbf{1.83} & 0.98/\textbf{1.45} \\
Science vs Art & 1.38/\textbf{1.66} & 0.92/\textbf{1.44} \\
Physical vs. Mental condition & 0.42/\textbf{0.64} & -0.12/\textbf{-0.77} \\
\midrule
\textbf{WAT} & \textbf{0.53}/0.41 & \textbf{0.46}/0.31 \\ 
\midrule
\textbf{SSSB} & & \\
black (race) & \textbf{5.36}/4.64 & 5.40/\textbf{5.67} \\
black (colour) & \textbf{5.36}/1.64 & \textbf{5.40}/4.83 \\
\cdashlinelr{1-3}
nationality & \textbf{7.78}/7.01 & \textbf{6.94}/5.75 \\
language & 7.78/\textbf{8.23} & 6.94/\textbf{7.38} \\
\cdashlinelr{1-3}
noun & 0.34/\textbf{0.39} & 0.09/\textbf{0.16} \\
verb & \textbf{0.34}/0.26 & \textbf{0.09}/0.06 \\
\bottomrule
\end{tabular}
}
\caption{Bias in LMMS and ARES Static Sense Embeddings.
In each row, between sense-insensitive word embeddings and sense embeddings, the larger deviation from 0 is shown in bold.
All results on WEAT are statistically signiciant ($p<0.05$) according to \eqref{eq:effect}.}
\vspace{-3mm}
\label{tbl:staic}
\end{table}

In the bias scores reported for \emph{nationality} vs.~\emph{language} senses, we find that \emph{nationality} obtains higher biases at word-level, while \emph{language} at the sense-level in both LMMS and ARES.
Unlike \emph{black}, where the two senses (colour vs.~race) are distinct, the two senses \emph{nationality} and \emph{language} are much closer because in many cases (e.g. Japanese, Chinese, Spanish, French etc.) languages and nationalities are used interchangeably to refer to the same set of entities.
Interestingly, the \emph{language} sense is assigned a slightly higher bias score than the \emph{nationality} sense in both LMMS and ARES sense embeddings.
Moreover, we see that the difference between the bias scores for the two senses in \emph{colour} vs.~\emph{race} (for black) as well as \emph{nationality} vs.~\emph{language} is more in LMMS compared to that in ARES sense embeddings.

\begin{figure}[t]
\centering
\includegraphics[width=0.48\textwidth]{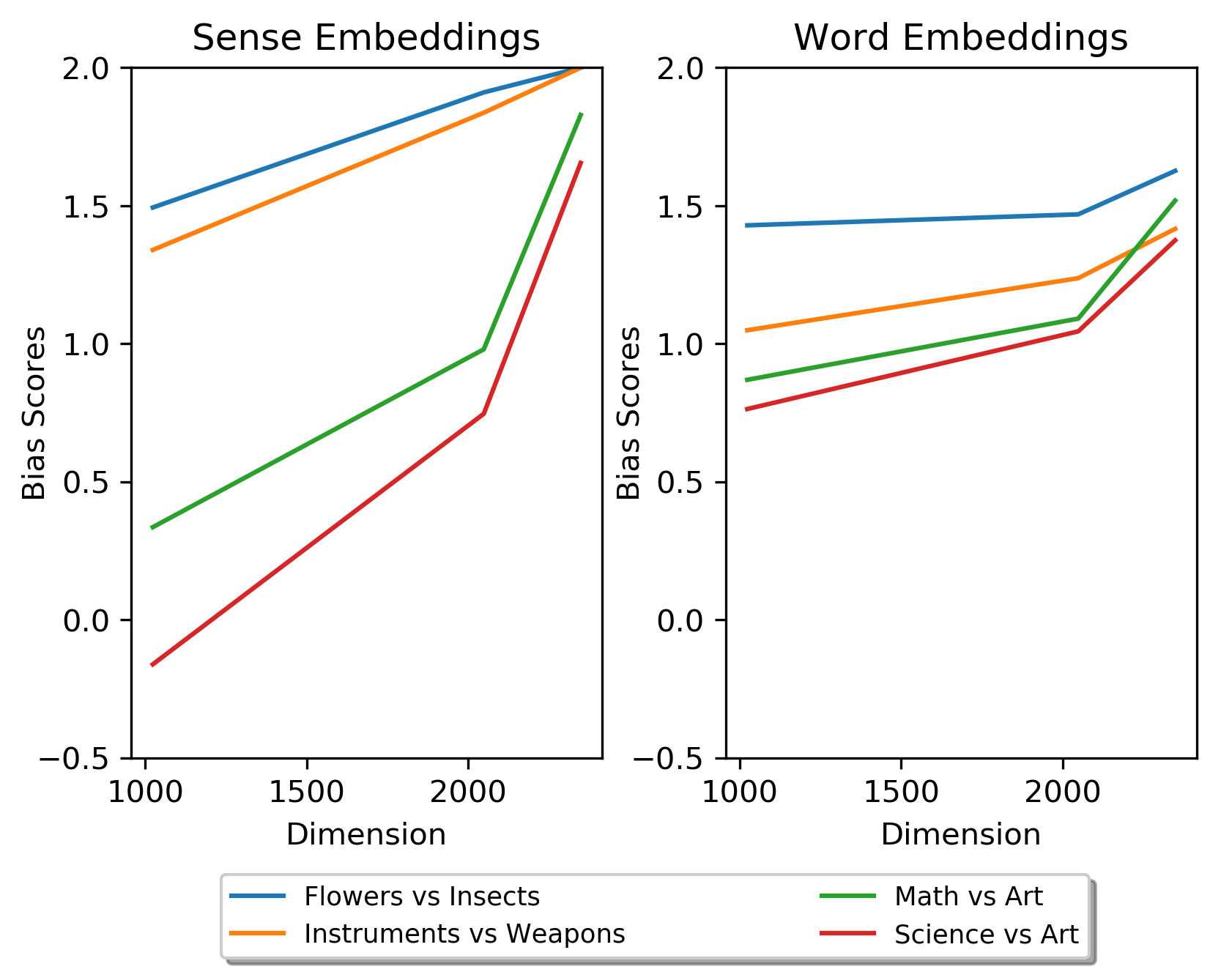}
\caption{Effect of the dimensionality of sense embeddings (LMMS) and word embeddings (LMMS-average).}
\label{fig:dimensionality-weat}
\vspace{-3mm}
\end{figure}

Between noun vs.~verb senses of occupations, we see a higher gender bias for the noun sense than the verb sense in both LMMS and ARES sense embeddings.
This agrees with the intuition that gender biases exist with respect to occupations and not so much regarding what actions/tasks are carried out by the persons holding those occupations.
Compared to the word embeddings, there is a higher bias for the sense embeddings in the noun sense for both LMMS and ARES.
This trend is reversed for the verb sense where we see higher bias scores for the word embeddings than the corresponding sense embeddings in both LMMS and ARES.
Considering that gender is associated with the noun than verb sense of occupations in English,
this shows that there are hidden gender biases that are not visible at the word-level but become more apparent at the sense-level.
This is an important factor to consider when evaluating gender biases in word embeddings, which has been largely ignored thus far in prior work.

To study the relationship between the dimensionality of the embedding space and the social biases it encodes, we compare 1024, 2048 and 2348 dimensional LMMS static sense embeddings and their corresponding word embeddings (computed using \eqref{eq:word-embeddings-average}) on the WEAT dataset in \autoref{fig:dimensionality-weat}.
We see that all types of social biases increase with the dimensionality for both word and sense embeddings.
This is in agreement with~\newcite{silva-etal-2021-towards} who also reported that increasing model capacity in contextualised word embeddings does not necessarily remove their unfair social biases.
Moreover, in higher dimensionalities sense embeddings show a higher degree of social biases than the corresponding (sense-insensitive) word embeddings.


\subsection{Bias in Contextualised Embeddings}
\label{sec:context}

To evaluate biases in contextualised sense embeddings, we use SenseBERT\footnote{\url{https://github.com/AI21Labs/sense-bert}}~\cite{levine-etal-2020-sensebert}, which is a fine-tuned version of BERT\footnote{\url{https://github.com/huggingface/transformers}}~\cite{devlin2019bert} to predict supersenses in the WordNet. 
For both BERT and SenseBERT, we use base and large pretrained models of dimensionalities respectively $768$ and $1024$.
Using AUL, we compare biases in BERT and SenseBERT using SSSB, CrowS-Pairs and StereoSet\footnote{We use only intrasentence test cases in StereoSet.} datasets.
Note that unlike SSSB, CrowS-Pairs and StereoSet \emph{do not} annotate for word senses, hence cannot be used to evaluate sense-specific biases.

\begin{table}[t]
\centering
\small
\begin{tabular}{lcc}
\toprule
        & base & large \\
\cmidrule(lr){2-2} \cmidrule(lr){3-3}
Dataset & BERT/SenseBERT & BERT/SenseBERT \\
\midrule
\textbf{CrowS-Pairs} & \textbf{-1.66}/0.99 & \textbf{-3.58}/2.45 \\
\textbf{StereoSet} & -1.09/\textbf{8.31} & -1.47/\textbf{6.51} \\
\midrule
\textbf{SSSB} & & \\
race & 10.19/\textbf{14.81} & \textbf{-17.59}/0.00 \\
colour & \textbf{-6.64}/-2.96 & -8.88/\textbf{9.84} \\
\cdashlinelr{1-3}
nationality & 5.79/\textbf{15.34} & 4.28/\textbf{8.10} \\
language & -0.17/\textbf{-2.95} & \textbf{6.25}/-3.82 \\
\cdashline{1-3}
noun  & 10.42/\textbf{14.06} & \textbf{3.13}/\textbf{3.13} \\
verb  & \textbf{12.89}/-3.74 & 0.22/\textbf{-15.44} \\
\bottomrule
\end{tabular}
\caption{Bias in BERT and SenseBERT contextualised word/sense embeddings.
In each row, between the AUL bias scores for the word vs.~sense embeddings, the larger deviation from 0 is shown in bold.
}
\label{tbl:context}
\end{table}

\autoref{tbl:context} compares the social biases in contextualised word/sense embeddings.
For both base and large versions, we see that in CrowS-Pairs, BERT to be more biased than SenseBERT, whereas the opposite is true in StereoSet.
Among the nine bias types included in CrowS-Pairs, \emph{gender} bias related test instances are the second most frequent following \emph{racial} bias.
On the other hand, gender bias related examples are relatively less frequent in StereoSet 
(cf. gender is the third most frequent bias type with 40 instances among the four bias types in StereoSet following \emph{race} with 149 instances and \emph{profession} with 120 instances out of the total 321 intrasentence instances).
This difference in the composition of bias types explains why the bias score of BERT is higher in CrowS-Pairs, while the same is higher for SenseBERT in StereoSet.

In SSSB, in 8 out of the 12 cases SenseBERT demonstrates equal or higher absolute bias scores than BERT.
This result shows that even in situations where no biases are observed at the word-level, there can still be significant degrees of biases at the sense-level.
In some cases (e.g. \emph{verb} sense in base models and \emph{colour}, \emph{language} and \emph{verb} senses for the large models), we see that the direction of the bias is opposite between BERT and SenseBERT.
Moreover, comparing with the corresponding bias scores reported by the static word/sense embeddings in \autoref{tbl:staic}, we see higher bias scores reported by the contextualised word/sense embeddings in \autoref{tbl:context}.
Therefore, we recommend future work studying social biases to consider not only word embedding models but also sense embedding models.


\section{Gender Biases in SSSB}
\label{sup:sec:noun-verb}

In this section, we further study the gender-related biases in static and contextualised word and sense embeddings using the noun vs.~verb sense instances (described in \autoref{sec:noun-vs-verb}) in the SSSB dataset.
To evaluate the gender bias in contextualised word/sense embeddings we use AUL on test sentences in SSSB noun vs.~verb category.
To evaluate the gender bias in static embeddings, we follow \newcite{Tolga:NIPS:2016} and use
the cosine similarity between (a) the static word/sense embedding of the occupation corresponding to its noun or verb sense and (b) the gender directional vector $\vec{g}$, given by \eqref{eq:gender}:
\begin{align}
\label{eq:gender}
     \vec{g} = \frac{1}{|\mathcal{C}|}\sum_{(m,f) \in \mathcal{C}} (\vec{m} - \vec{f})
\end{align}

Here, $(m, f)$ are male-female word pairs used by \citet{kaneko-bollegala-2019-gender} such as (\emph{he}, \emph{she}) and $\vec{m}$ and $\vec{f}$ respectively denote their word embeddings.
Corresponding sense-insensitive word embeddings are computed for the 2048 dimensional LMMS embeddings using \eqref{eq:word-embeddings-average}. 

\begin{table*}[t]
    \centering
    \begin{adjustbox}{width=\linewidth}
    \begin{tabular}{l ccc ccc}
        & \multicolumn{3}{c}{BERT} & \multicolumn{3}{c}{SenseBERT} \\
       \textcolor{red}{stereo}/\textcolor{blue}{anti-stereo} sentences & stereo & anti & diff & stereo & anti & diff \\ \toprule
        \textcolor{red}{he}/\textcolor{blue}{she} is a strong nurse & -0.45 & -0.67 & 0.22 & -15.71 & -16.64 & 0.93 \\
       \textcolor{red}{he}/\textcolor{blue}{she} is a professional nurse & -0.73 & -0.85 & 0.11 & -16.53 & 16.81 & 0.27 \\
        As a \textcolor{red}{mother}/\textcolor{blue}{father} of five, \textcolor{red}{she}/\textcolor{blue}{he} carefully nurse all of \textcolor{red}{her}/\textcolor{blue}{his} children & -0.16 & -0.15 & -0.01 & -18.07 & -18.24 & 0.18 \\
        \textcolor{red}{she}/\textcolor{blue}{he} made milk \textcolor{red}{herself}/\textcolor{blue}{himself} to nurse the crying baby & -0.77 & -0.14 & -0.63 & -15.85 & -17.80 & 1.96 \\\bottomrule
    \end{tabular}
    \end{adjustbox}
    \caption{Pseudo log-likelihood scores computed using Eq.~\eqref{eq:AUL} for \textcolor{red}{stereo} and \textcolor{blue}{anti-stereo} sentences (shown together due to space limitations) using BERT-base and SenseBERT-base models.
    Here, diff = \textcolor{red}{stereo} - \textcolor{blue}{anti}.}
    \label{tbl:examples}
\end{table*}

\begin{figure}[t]
\centering
\includegraphics[width=0.45\textwidth]{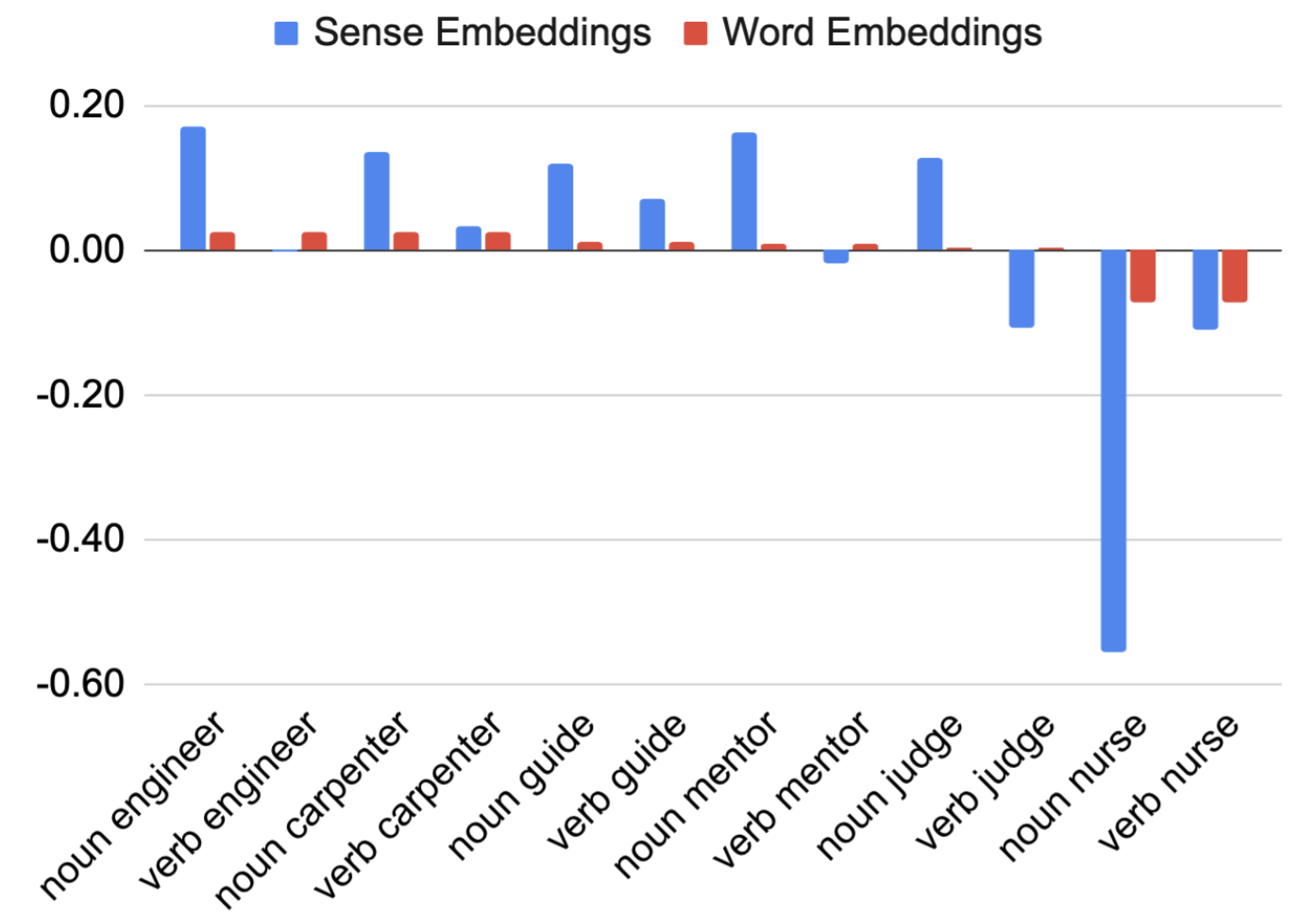}
\caption{Gender biases found in the 2048-dimensional LMMS static sense embeddings and corresponding word embeddings computed using \eqref{eq:word-embeddings-average}. 
Positive and negative cosine similarity scores with the gender directional vector (computed using \eqref{eq:gender}) represent biases towards respectively the \emph{male} and \emph{female} genders.}
\label{fig:static-gender}
\end{figure}

\autoref{fig:static-gender} shows the gender biases in LMMS embeddings.
Because static word embeddings are not sense-sensitive, they report the same bias scores for both noun and verb senses for each occupation.
For all noun senses, we see positive (male) biases, except for \emph{nurse}, which is strongly female-biased.
Moreover, compared to the noun senses, the verb senses of LMMS are relatively less gender biased.
This agrees with the intuition that occupations and not actions associated with those occupations are related to gender, hence can encode social biases.
Overall, we see stronger biases in sense embeddings than in the word embeddings.

\begin{figure}[t]
\centering
\includegraphics[width=0.48\textwidth]{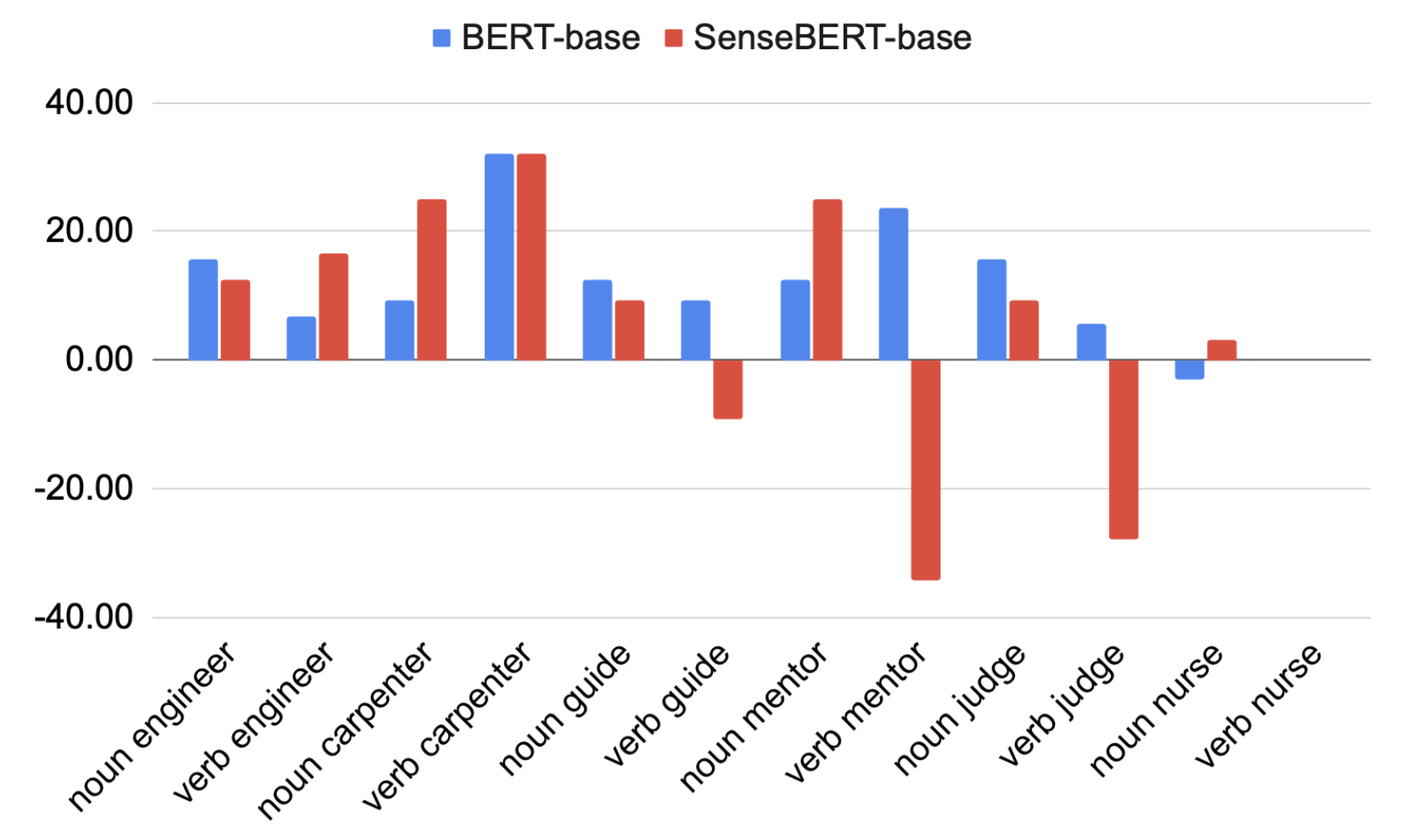}
\caption{Gender biases found in 768-dimensional BERT-base and SenseBERT-base contextualised embeddings. 
Positive and negative AUL scores represent bias towards respectively the stereotypical and anti-stereotypical sentences.}
\label{fig:contextual-gender}
\end{figure}

\autoref{fig:contextual-gender} shows the gender biases in BERT/SenseBERT embeddings.
Here again, we see that for all noun senses there are high stereotypical biases in both BERT and SenseBERT embeddings, except for \emph{nurse} where BERT is slightly anti-stereotypically biased whereas SenseBERT shows a similar in magnitude but a stereotypical bias.
Recall that \emph{nurse} is stereotypically associated with the female gender, whereas other occupations are predominantly associated with males, which is reflected in the AUL scores here.

Despite being not fine-tuned on word senses, BERT shows different bias scores for noun/verb senses, showing its ability to capture sense-related information via contexts.
The verb sense embeddings of SenseBERT of \emph{guide}, \emph{mentor} and \emph{judge} are anti-stereotypical, while the corresponding BERT embeddings are stereotypical.
This shows that contextualised word and sense embeddings can differ in both magnitude as well as direction of the bias.
Considering that SenseBERT is a fine-tuned version of BERT for a specific downstream NLP task (i.e. super-sense tagging), one must not blindly assume that an unbiased MLM to remain as such when fine-tuned on downstream tasks.
\emph{How social biases in word/sense embeddings change when used in downstream tasks} is an important research problem in its own right, which is beyond the scope of this paper.


A qualitative analysis is given in \autoref{tbl:examples} where the top-two sentences selected from SSSB express the noun sense of \emph{nurse}, whereas the bottom-two setences express its verb sense.
From \autoref{tbl:examples}, we see that SenseBERT has a higher preference (indicated by the high pseudo log-likelihood scores) for stereotypical examples than BERT over anti-stereotypical ones (indicated by the higher diff values).

\section{Conclusion}

We evaluated social biases in sense embeddings by extending existing word-level bias evaluation datasets (WEAT, WAT) and by creating a novel sense-specific contextualised dataset (SSSB).
Our experiments show that sense embeddings are also socially biased similar to word embeddings. 
Extending the analysis beyond English and developing debiasing methods for sense embedding are identified as important future research directions.

\section{Ethical Considerations}

In this paper we considered the relatively underexplored aspect of social biases in pretrained sense embeddings. 
We created a new dataset for this purpose, which we name the Sense-Sensitive Social Bias (SSSB) dataset.
The dataset we create is of a sensitive nature.
We have included various sentences that express stereotypical biases associated with different senses of words in this dataset.
We specifically considered three types of social biases in SSSB: (a) racial biases associated with a nationality as opposed to a language (e.g. \emph{Chinese people are cunning}, \emph{Chinese language is difficult}, etc.),
(b) racial biases associated with the word \emph{black} as opposed to its sense as a colour (e.g. \emph{Black people are arrogant}, \emph{Black dress is beautiful}, etc.) and
(c) gender-related biases associated with occupations used as nouns as opposed to verbs (e.g. \emph{She was a careless nurse}, \emph{He was not able to nurse the crying baby}, etc.).
As seen from the above-mentioned examples, by design, SSSB contains many offensive, stereotypical examples. 
It is intended to facilitate evaluation of social biases in sense embeddings and is publicly released for this purpose only.
We argue that SSSB should not be used to train sense embeddings. 
The motivation behind creating SSSB is to measure social biases so that we can make more progress towards debiasing them in the future. 
However, training on this data would defeat this purpose.

It is impossible to cover all types of social biases related to word senses in any single dataset.
For example, the stereotypical association of a disadvantaged group with a positive attribute (e.g. \emph{All Chinese students are good at studying}) can also raise unfairly high expectations for the members in that group and cause pressure to hold upto those stereotypes. Such \emph{positive biases} are not well covered by any of the existing bias evaluation datasets, including the one we annotate in this work.

Given that our dataset is generated from a handful of manually written templates, it is far from complete.
Moreover, the templates reflect the cultural and social norms of the annotators from a US-centric viewpoint. 
Therefore, SSSB should not be considered as an ultimate test for biases in sense embeddings.
Simply because a sense embedding does not show any social biases on SSSB according to the evaluation metrics we use in this paper \emph{does not} mean that it would be appropriate to deploy it in downstream NLP applications that require sense embeddings.
In particular, task-specific fine-tuning of even bias-free embeddings can result in novel unfair biases from creeping in.

Last but not least we state that the study conducted in this paper has been limited to the English language and represent social norms held by the annotators.
Moreover, our gender-bias evaluation is limited to binary (male vs. female) genders and racial-bias evaluation is limited to Black as a race. 
Extending the categories will be important and necessary future research directions.

\bibliography{Unbiased,sense}
\bibliographystyle{acl_natbib}

\end{document}